\documentclass[runningheads]{llncs}

\usepackage{eccv}

\usepackage{eccvabbrv}

\usepackage{graphicx}
\usepackage{booktabs}
\usepackage{multirow}
\usepackage{mathtools}

\usepackage[accsupp]{axessibility}  
\usepackage{hyperref}

\usepackage{orcidlink}

\begin{document}


\title{Video Face Re-Aging: Toward Temporally Consistent Face Re-Aging} 

\author{Abdul Muqeet$^*$ \and
Kyuchul Lee$^*$ \and
Bumsoo Kim \and
Yohan Hong \and
Hyungrae Lee \and
Woonggon Kim \and
KwangHee Lee}

\authorrunning{A Muqeet et al.}

\institute{VIVE STUDIOS \\
\email\{amuqeet, lkc880425, bskim, hongsusoo, hyunlee, wgkim, lucas\}@vivestudios.com}
 
{
  \renewcommand{\thefootnote}%
    {\fnsymbol{footnote}}
  \footnotetext[1]{  Equal contribution.} 
}

\maketitle

\begin{abstract}
  Video face re-aging deals with altering the apparent age of a person to the target age in videos. This problem is challenging due to the lack of paired video datasets maintaining temporal consistency in identity and age. Most re-aging methods process each image individually without considering the temporal consistency of videos. While some existing works address the issue of temporal coherence through video facial attribute manipulation in latent space, they often fail to deliver satisfactory performance in age transformation. To tackle the issues, we propose (1) a novel synthetic video dataset that features subjects across a diverse range of age groups; (2) a baseline architecture designed to validate the effectiveness of our proposed dataset, and (3) the development of novel metrics tailored explicitly for evaluating the temporal consistency of video re-aging techniques. Our comprehensive experiments on public datasets, including VFHQ and CelebA-HQ, show that our method outperforms existing approaches in age transformation accuracy and temporal consistency. Notably, in user studies, our method was preferred for temporal consistency by 48.1\% of participants for the older direction and by 39.3\% for the younger direction.

\end{abstract}

\section{Introduction}
\label{sec:introduction}

Video Face re-aging aims to transform the apparent age in facial videos while ensuring the temporal consistency in both age and identity. This field holds significance relevance across diverse domains, including computer graphics, forensics, entertainment, and advertising. Despite the extensive research conducted in this domain, the challenge remains largely unexplored when it comes to videos. One of the remaining challenges is that existing image-based methods yield inconsistent identities when applied to videos or consecutive frames featuring varying expressions, viewpoints, and lighting conditions.

Recent studies \cite{alaluf2022third,tzaban2022stitch} have leveraged StyleGAN-based \cite{karras2019style} frameworks to develop techniques for manipulating facial attributes in videos, aiming for greater attribute consistency. However, these approaches frequently fall short in precisely executing the desired age transformations. They commonly rely on assuming a linear path in the latent space for explicit age control, which is not always accurate. As a result, noticeable artifacts frequently emerge in the output when dealing with input data featuring significant age gaps.

Lately, \cite{zoss2022production} have utilized the synthetic images labeled with existing re-aging techniques \cite{alaluf2021only} to curate a paired dataset for re-aging. Their study shows that supervised training on synthetic dataset yields favorable outcomes for still images. However, it is well-known that models trained on static images often suffer from temporal inconsistencies.

These observations suggest that training face re-aging on video datasets is beneficial for addressing temporal consistency issues in video re-aging tasks. Therefore, we propose a pipeline to generate a synthetic video dataset. This dataset comprises paired data for supervised training, consisting of various ages, poses, and expressions. Creating this dataset involves three major steps.

Firstly, we utilize StyleGAN to synthesize a face image of a specific individual. Then we apply existing re-aging method SAM \cite{alaluf2021only} to generate images of the same person with varying ages (Sec. \ref{sec:data_generation}). Next, we generate key frames which consist of various poses and expressions of that individual (Sec. \ref{sec:keyframes_generation}). Lastly, we produce a natural and continuous motion through key frames (Sec. \ref{sec:motion_generation}).

As a result of our pipeline, we obtain paired videos that feature individuals of different ages and in various poses and expressions. For the first time, this has enabled us to construct a video re-aging dataset for supervised training. Our work can be viewed as an extension of the existing research \cite{zoss2022production}, which previously introduced a synthetic image dataset for re-aging.

 In addition to the dataset, we introduce a baseline architecture designed to utilize the temporal coherence inherent in our proposed video dataset. This architecture primarily comprises recurrent blocks, employing a fusion-based approach that leverages concatenated inputs to exploit temporal consistency. Drawing inspiration from seminal works in video generation \cite{clark2019adversarial,saito2020train,tulyakov2018mocogan}, we incorporate a video discriminator equipped with 3D convolutional layers to ensure both realism and natural motion in the generated videos.
 
Recognizing that existing aging metrics are not well-suited for video-based methods, we address this gap by developing novel metrics to assess temporal continuity in video-based re-aging. Through extensive experiments, we demonstrate that our video-based architecture produces remarkable results and outperforms existing state-of-the-art methods across various public datasets.

\begin{figure*}[ht]
  \centering
   \includegraphics [width=\textwidth]{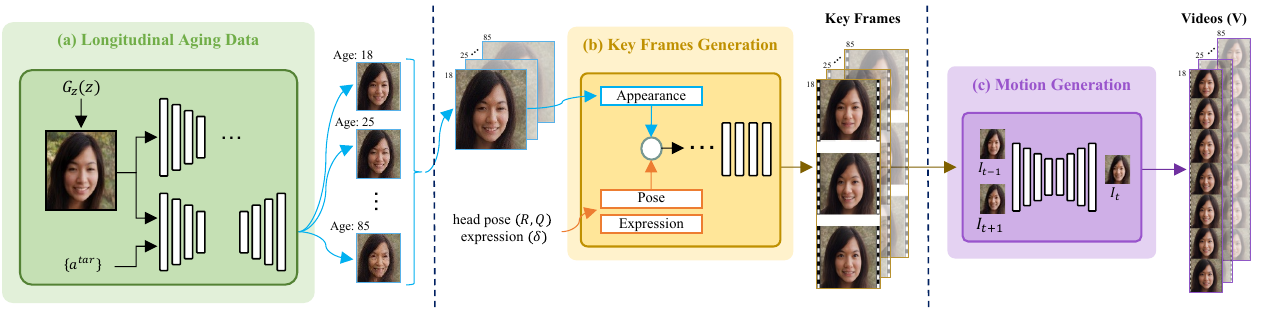} 
  \caption{Our proposed pipeline to construct the video dataset for re-aging. Firstly, high-resolution synthetic facial images are created using StyleGAN \cite{karras2019style} Subsequently, images of individuals at different target ages are generated using SAM \cite{alaluf2021only} for age transformation. Next, key frames are produced by employing OSFV, which alters the pose and expression of these synthetic images. This is achieved without relying on driving images, instead using random values for rotation, translation, and expression keypoints. Finally, motion is added to these key frames using FILM \cite{reda2022film}, creating smooth and high-fidelity motion videos of subjects at different ages.}
  \label{building_dataset}
\end{figure*}

Given the challenges and limitations of current methods, our work introduces several contributions in video re-aging as follows.

\begin{enumerate}

\item
    For the first time, we introduce a pipeline designed to generate a synthetic video dataset specifically for video re-aging. This dataset features videos of individuals across various ages, poses, and expressions.

\item
    We present a baseline network architecture custom-designed for our synthesized video dataset. Our generator is built upon a combination of recurrent blocks and U-Net modules, and it utilizes both 2D image-based and 3D video-based conditional discriminators. 
    
\item
    We propose novel metrics tailored to assess the temporal consistency of re-aging methods. These include Temporal Regional Wrinkle Consistency (TRWC) and Temporally Age Preservation. These metrics provide a robust framework for evaluating the quality of age transformations over time.
\end{enumerate}

\section{Related Works}
\label{sec:related_work}

\subsection{Face Image Re-Aging}

The study in \cite{antipov2017face} pioneered the use of a conditional GAN for face aging. Subsequently, several influential works such as \cite{antipov2017boosting,wang2018face,or2020lifespan, li2021continuous,yao2021high,kim2024toonaging} emerged, expanding on this concept. Similar to \cite{li2021continuous}, SAM \cite{alaluf2021only} emphasizes continuous age progression. SAM is a StyleGAN-based model \cite{karras2019style} 
 capable of generating high-resolution images. In contrast to other methods, it does not use an age classifier to estimate the input age. FRAN \cite{zoss2022production} 
trains a simple encoder-decoder network in a supervised manner with a generated synthetic dataset. Rather than adopting age classifiers or embedding, it extends the input to a 5-channel image, including two binary masks for input and target ages. CUSP \cite{gomez2022custom} introduces style and content encoders to disentangle the style and content of the input. This approach also incorporates the GB algorithm \cite{springenberg2014striving} within the CUSP module, ensuring that only age-relevant features are processed. AgeTransGAN \cite{hsu2022agetransgan} disentangles the encoded image into identity and age components with two independent modules. PADA \cite{li2023pluralistic} and FADING \cite{chen2023fading} adopt a text-driven approach and integrates the pre-trained CLIP \cite{radford2021learning} and diffusion models. However, these works only focus on images, and directly applying these methods on individual frames does not consider temporal consistency, affecting the quality of re-aging quality over time.

\subsection{Face Video Re-Aging}

Most video face re-aging methods transform the face by manipulating the age in latent space, except \cite{duong2019automatic} that proposes a reinforcement learning method for the sequence of video frames. \cite{yao2021latent} proposes editing the vectors in the StyleGAN latent space with various pre-processing steps that are independently applied to every single frame of an input video. \cite{tzaban2022stitch} identify the inconsistencies in PTI \cite{roich2022pivotal} and suggest e4e \cite{tov2021designing} encoder with it for finding the pivots. Video editing methods often crop faces as a prepossessing step. To overcome this problem, \cite{yang2023styleganex} addresses this issue by proposing changes in the initial StyleGAN layers to overcome the cropping problem. \cite{kim2023diffusion} proposes a diffusion-based editing approach by using \cite{preechakul2022diffusion} that disentangles the video into time-dependent features (such as motion) that are applied to each frame and time-independent features (such as identity) which are shared across all the frames. While improving temporal consistency over image-based methods, they still struggle to accurately transform faces to the target age.

\subsection{Re-Aging Dataset} Video face re-aging presents significant challenges, primarily due to the lack of dedicated video datasets. Existing image-based face re-aging datasets are typically labeled either automatically using an age classifier or manually via crowdsourcing \cite{zhang2022learning,liu2021learning,karras2019style,rothe2018deep}. However, these datasets do not provide paired ages for supervised training. \cite{zheng2017cross} proposed a technique to generate a synthetic dataset that is labelled through SAM \cite{alaluf2021only}. \cite{duong2019automatic} also labelled a video dataset, which remains private.

\section{Video Face Re-Aging Framework}
\label{sec:method}

We first describe the pipeline of synthesizing the proposed video dataset. This is followed by introducing our baseline architecture and loss functions. Lastly, we present novel metrics specifically designed for video re-aging methods to quantify their re-aging performance over time.

\setcounter{secnumdepth}{5}
\subsection{Re-Aging Video Dataset}
\label{sec:dataset}

\subsubsection{Longitudinal Aging Data}
\label{sec:data_generation}
Creating a high-quality image re-aging dataset is a crucial in our pipeline because it directly influences the quality of our subsequent video re-aging dataset. However, obtaining these paired image datasets is a challenging task. To overcome this limitation, we turn to insights from \cite{zoss2022production}, which show that training on synthetic datasets can yield realistic results on real images. For instance, the neural network can learn how wrinkles change from the synthetic images and apply this knowledge to real images. We refer to these learned changes as delta images $D_t$, as illustrated in Fig. \ref{vfran_architecture}.

Firstly, we leverage StyleGAN \cite{karras2020analyzing} which takes a random noise as input and generate high-resolution synthetic image. Then we utilize SAM \cite{alaluf2021only} that manipulates the latent vector with StyleGAN for age transformation. The intuition behind choosing SAM over other existing methods is its re-aging performance in terms of age error, that is also evident through our experiments. Using this approach, we construct a synthetic facial image dataset:

\begin{equation}
I^{tar} = SAM(G_z(z); a^{tar}) 
\end{equation}

Given a random noise $z$, StyleGAN $G_z(\cdot)$ produces a sample image, which is then used as input for SAM \cite{alaluf2021only} along with a target age $a^{tar}$. As a result we get an image $I^{tar}$ with apparent age $a^{tar}$ as shown in Fig. \ref{building_dataset} (a). Readers can refer to \cite{zoss2022production} for more details about image-level re-aging as this process is greatly inspired by \cite{zoss2022production}.

\subsubsection{Key Frames Generation} 
\label{sec:keyframes_generation}

The subsequent step in the outlined pipeline involves key frames generation. This key frames capture specific snapshots or moments in videos to enable seamless transitions between sequences. The challenge lies in acquiring diverse facial images to adequately encapsulate the video's dynamics, including various poses and expressions. To address this, we utilize a recent face reenactment technique \cite{wang2021one} that modifies the pose and expression of a source image based on a driving image. Incorporating this method into our pipeline allows us to generate multiple images of an individual with varying poses and expressions. These resulting images serve as key frames.

In this study, we employ the off-the-shelf model OSFV\footnote{Note that we trained OSFV on VFHQ \cite{xie2022vfhq} dataset for a resolution of 512$\times$512. The training process on VFHQ dataset requires 30 days on 8 A100 GPUs. Additional details are provided in the supplementary materials.}\cite{wang2021one} to produce synthetic key frames using our image dataset. Our approach diverges from the original methodology in that we rely solely on source images from our dataset. Instead of utilizing driving images, which can be challenging to gather, we employ random values for the rotation matrix $R$, translation matrix $Q$, and expression keypoints $\delta$ to generate various poses and expressions as shown in Fig. \ref{building_dataset} (b). We can write this process as follows:
 
\begin{equation}
K = G_{kp}(I^{tar}, R, Q, \delta ), 
\end{equation}

\noindent where $K$ represents generated key frames and $G_{kp}$ denotes key frames generator. We generate eight different key frames through this method. While it is also possible to repeat this procedure to create a motion video, but we have observed a degradation of the quality in the resultant videos. Therefore, we come up with an alternative approach to address this problem in next section.

\subsubsection{Motion Generation} 
\label{sec:motion_generation}
The final step of our pipeline is motion generation. We leverage the recent work in frame interpolation methods to ensure smooth and high-fidelity motion. Specifically, we employ the method presented in \cite{reda2022film} to the eight key frames generated in the previous step by recursively generating intermediate frames between them. This iterative process is executed for two consecutive frames and is repeated for every subject across all ages as follows:

\begin{equation}
{I}_{t} = FI(I_{t-1}, I_{t+1}),
\end{equation}

\noindent where $FI$ is motion generation network \cite{reda2022film} and $I_t$ is the frame at $t$ time-step. As a result, we obtain smooth and high-fidelity paired videos for every subject of different ages as shown in Fig. \ref{building_dataset} (c).

\begin{figure}[ht]
  \centering
  \includegraphics[width=\linewidth]{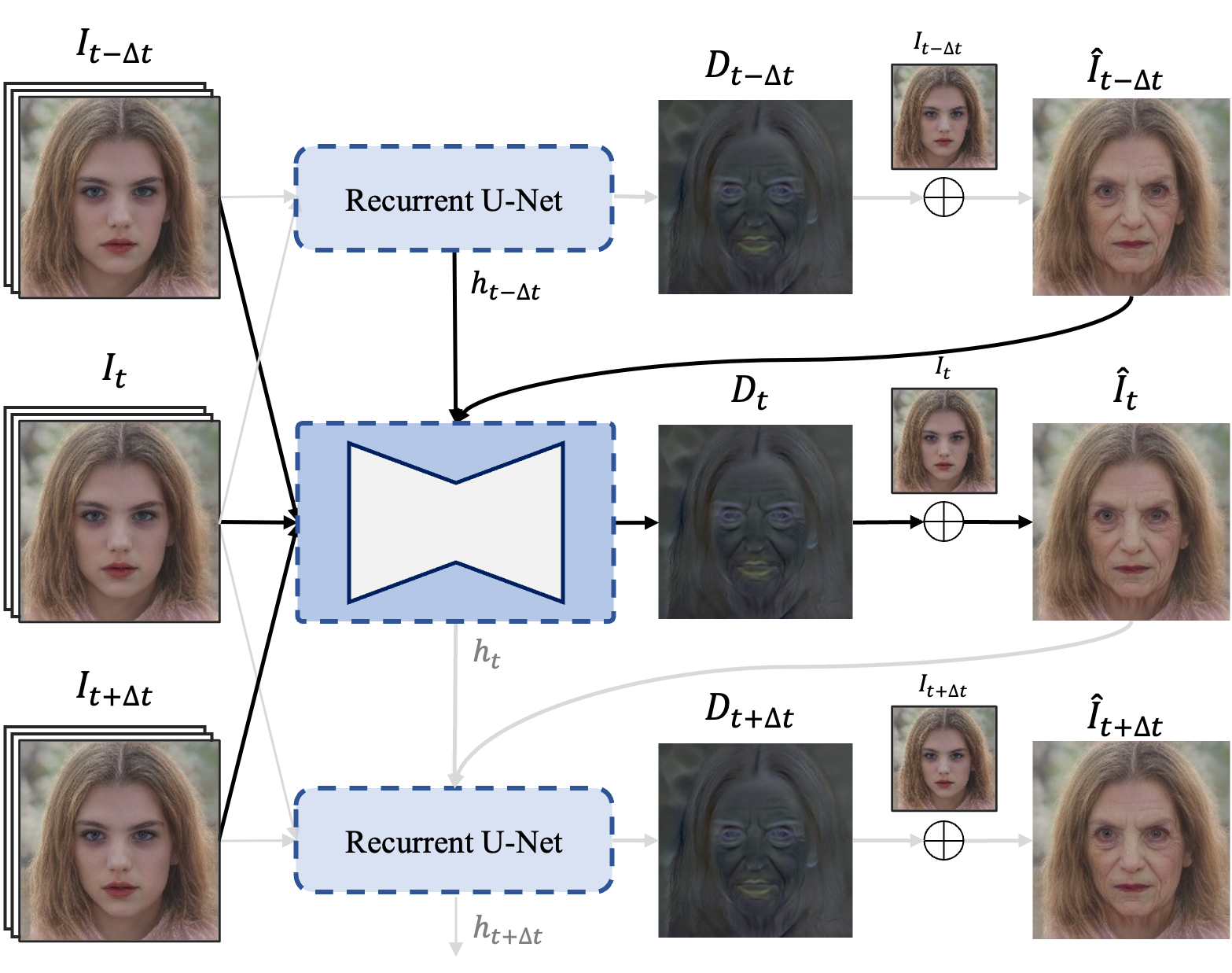}
  \caption{Overview of our generator for video re-aging.}\label{vfran_architecture}
\end{figure}

\setcounter{secnumdepth}{2}

\subsection{Network Architecture}

\subsubsection{Generator} To fully utilize the advantage of our novel video re-aging dataset, we carefully design the generation scheme to accurately transform faces  over time. Following the approach in \cite{zoss2022production}, we adopt an alternative method for incorporating input and target ages into our model structure. Recent methods \cite{gomez2022custom,hsu2022agetransgan} tend to use pre-trained age classifiers to estimate the input age. We concatenate the input frame $I_t$ at $t$ time-step with two spatial masks (one for input age and one for target age) over channel dimensions. These two masks contain constant values representing input and output ages, ranging from 0 to 100 normalized between 0 and 1. This results in a 5-channel masked frame $I^{mask}_{t}$. Mathematically,

\begin{equation}    
   I_t^{mask} = [I_t,M^{inp},M^{tar}], 
\end{equation}

\noindent where $M^{inp}$ and $M^{tar}$ are spatial masks for input and target ages. $[\cdot]$ denotes the channel-wise concatenation operation. If $V^{mask}=\{I^{mask}_1,I^{mask}_2,\dots,I^{mask}_N\}$ refers to any input masked sequence $V^{mask}$ for $N$ total number of frames, our generator $G$ produces the output video $\hat{V}$.

\begin{equation}    
 \tilde{V} = G( V^{mask}) ,
\end{equation}

The proposed $G$ adopts a recursive scheme to consider the temporal information of videos, inspired by \cite{ wu2022animesr, tian2021good, zhu2023motionvideogan}, namely recurrent block function $RB(\cdot)$. We employ the U-Net architecture within the $RB$ function. Please refer to Sec. \ref{sec:Experimental} for the details of U-Net network architecture. This $RB$ stacks the multiple arguments. First, we concatenate the consecutive input frames $[I^{mask}_{t-\Delta t},I^{mask}_{t},I^{mask}_{t+\Delta t}]$. Here, ${I^{mask}_{t-\Delta t}}$ and ${I^{mask}_{t+\Delta t}}$ refers to the two adjacent frames of $I^{mask}_t$ with interval step $\Delta t$. We further concatenate the resulting stacked frames with previous hidden state $h_{t-\Delta t}$ and previous output frame $\hat I_{t-\Delta t}$. We formulate this overall process as follows:

\begin{equation} \label{eq:rb}
h_{t}, {D}_t = RB( [{I^{mask}_{t-\Delta t}}, {I^{mask}_{t}}, {I^{mask}_{t+\Delta t}}, \tilde{I}_{t-\Delta t}, h_{t-\Delta t}])
\end{equation}   

As a result, we obtain hidden state $h_t$ and delta image ${D}_t$. Once the delta image is attained, we can easily obtain output frame $\hat{I}_t$ through element-wise summation between input $I_{t}$ and $D_t$:

\begin{equation}    
  \hat{I}_t = {D}_t  + {I_{t}}. 
\end{equation}

We have now processed three frames to get the re-aged output of the middle frame. For the remaining consecutive frames, we can pass the hidden state $h_{t}$ and output frame $\hat{I}_t$ along with the next consecutive frames in Eq. \ref{eq:rb} for the next iteration. This process is repeated for all $N$ frames of the input video $V^{mask}$ to generate the output video $\hat{V}$, as illustrated in Fig. \ref{vfran_architecture}.

\subsubsection{Discriminator} 
In addition to the image discriminator, we introduce a video discriminator to assess the consistency of age-related high frequency details across consecutive frames. We observed that using only image discriminator produces inconsistent output, leading to flickering. The comparison is presented in our ablation studies.  The image discriminator employs a PatchGAN architecture \cite{pix2pix2017} to differentiate realistic images from synthetic ones, while for the video discriminator, we adopt a spatio-temporal convolutional network that utilizes 3D convolution layers, similar to the approach used in previous works such as \cite{vondrick2016generating} and \cite{tulyakov2018mocogan}. Specifically, we exploit the spatio-temporal information from consecutive generated frames in the form of delta images $D_t$ as inputs to ensure the preservation of high-level details over time, rather than relying on facial images.

\subsubsection{Loss Functions}

We use the similar loss function as \cite{zoss2022production} by including $L1$ loss, LPIPS loss \cite{zhang2018unreasonable}, and adversarial loss functions \cite{lim2017geometric}. We also use adversarial loss functions for image and video discriminators $\mathcal{L}_{adv, I}$ and $\mathcal{L}_{adv, V}$. Our total objective is obtained as:

\begin{align*}
\label{eq:loss}
  \mathcal{L} = \lambda_{L1}\mathcal{L}_{L1} (\hat{V}, V_{gt}) + \lambda_{adv,V}\mathcal{L}_{adv,V}( \hat{V}, M^{tar} ) + \\\ \lambda_{adv,I} \mathcal{L}_{adv,I}(\hat{V}, M^{tar}) + \lambda_{p}\mathcal{L}_{LPIPS} (\hat{V}, V_{gt}), 
\end{align*}

\noindent where $V_{gt}$ is the ground-truth video corresponding to the target age obtained from dataset building pipeline. Here, we set the lambda values as $\lambda_{L1}=1.0$, $\lambda_{adv,I}=0.025$, $\lambda_{adv,V}=0.025$, $\lambda_{p}=1.0$.

\subsection{Proposed Metrics}

Evaluating age transformation performance in a video context is both essential and challenging. According to the best of our knowledge, there is no specific metric to assess the temporal consistency of re-aging methods. Most existing evaluation metrics are designed for image-based re-aging methods as they measure the age transformation performance on individual images \cite{zoss2022production} or use metrics designed to evaluate the continuity of identity drift \cite{alaluf2022third}. Regarding this, we propose two metrics, TRWC and T-Age, capable of effectively assessing temporal continuity in terms of age in video-based re-aging task.

\vspace{-1.5em}
\subsubsection{TRWC}

\begin{figure}[h]
  \centering
  \includegraphics[width=\linewidth]{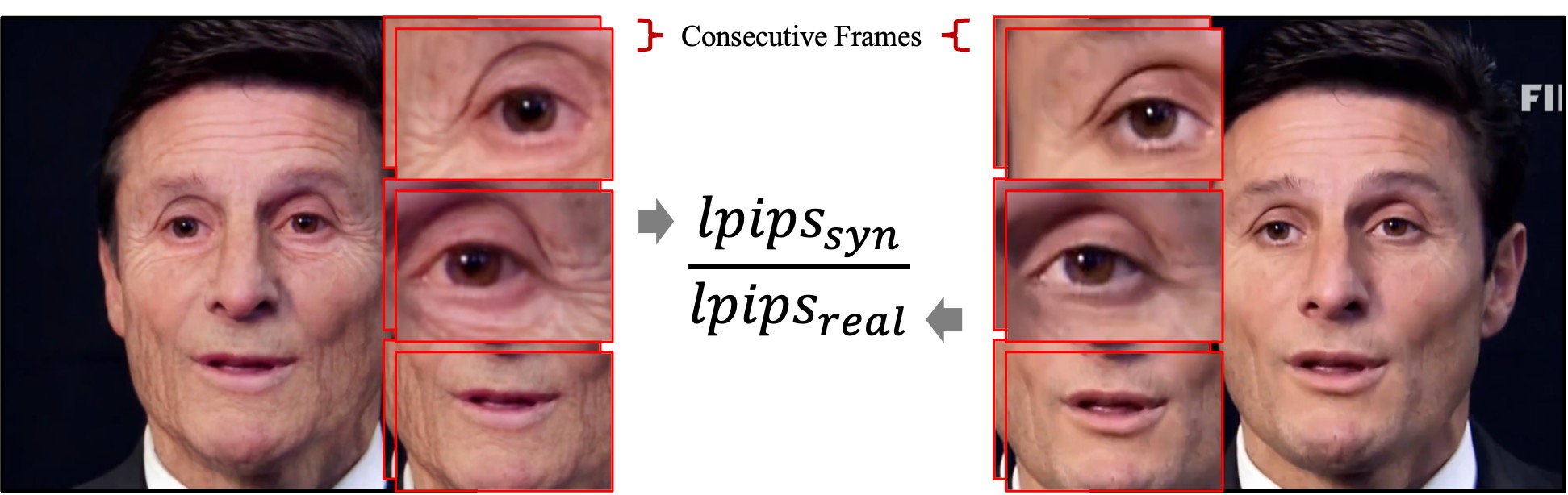}
  \caption{Conceptual overview of proposed TRWC.}
  \label{figure_lpips}
\end{figure}

Te evaluate the consistency of video re-aging methods, we need to consider the aging-related details in facial areas such as wrinkles where it matters. This is because these regions are most vulnerable and simultaneously important in terms of human perception. Therefore, we focus on aging-related facial areas, especially Crow's feet and Nasolabial folds, motivated from \cite{wang2021gfpgan, li2022learning}. Due to their active role in expressions and speech, these regions are highly susceptible to aging. We extract the region-of-interest (ROI) around these areas to observe changes over time. However, when analyzing perceptual differences between frames, factors such as facial angles and expressions can be sensitive. To address this, we calculate the LPIPS of the synthesized image and normalize it using the LPIPS of the real image, ensuring we consider differences in each image over time. Fig. \ref{figure_lpips} explains the conceptual schematic of TRWC. Mathematically, we define TRWC as follows.

\begin{equation}
    \begin{multlined}
    \text{TRWC}_{\Delta t} = \frac{1}{3(N-{\Delta t})} \\\sum_{r=1}^{3}\sum_{t=1}^{N-\Delta t}\frac{lpips(f^r_{roi}(\hat{I}_t), f^r_{roi}(\hat{I}_{t+{\Delta t}}))}{lpips(f^r_{roi}(I_t), f^r_{roi}(I_{t+{\Delta t}}))},
    \end{multlined}
    \label{eq: TRWC}
\end{equation}

\noindent where $N$ is the number of frames and ${\Delta t}$ is time interval between consecutive frames. The function $f^r_{roi}(\cdot)$ is ROI function that includes eyes and mouth obtained by facial landmark detector \cite{deng2018menpo}. Here, we consider only three ROI regions, left-eye, right-eye, and mouth, indexed by $r$. 

\subsubsection{T-Age}
Drawing inspiration from the TL-ID metric \cite{tzaban2022stitch}, which proposes metrics for identity consistency in videos, we introduce T-Age for video re-aging. T-Age measures the age difference between two adjacent frames using cosine similarity, utilizing an off-the-shelf age classifier \cite{rothe2015dex}. A lower T-Age value indicates a more consistent age representation across the frames.

\section{Experiments}

\subsection{Experiment Setup}

In this section, we show the superiority of our method by comparing with all existing state-of-the-art re-aging methods. This includes HRFAE \cite{yao2021high}, SAM \cite{alaluf2021only}, AgeTransGAN \cite{hsu2022agetransgan}, Diffusion AE \cite{preechakul2022diffusion}, STIT \cite{tzaban2022stitch}, StyleGANEX \cite{yang2023styleganex}, Diffusion VAE \cite{kim2023diffusion}, and FRAN \cite{zoss2022production}, totaling 8 methodologies. 

We have trained our methods on the proposed synthetic videos generated through our pipeline. For the test set, we choose the CelebV-HQ \cite{zhu2022celebvhq} and VFHQ dataset \cite{xie2022vfhq} as video test set. We consider three target age groups, $(18,25,35)$ with input age as ${85}$ for $Old \rightarrow Young$ task and three age groups, $(65,75,85)$ with input age as ${18}$ for $Young \rightarrow Old$ task. We provide the additional details in the supplementary materials (Sec. \ref{sec:Experimental}).

\subsection{Metrics}

We evaluate the performance of re-aging models based on their ability to transform to the target age while ensuring temporal consistency in the re-aged image. Given that we do not have access to ground-truth for real videos, we employ metrics that do not rely on ground-truth. Specifically, we use four metrics to evaluate the results.  We calculate mean absolute error (MAE) between the estimate ages, computed with the pre-trained age classifier to quantify the age transformation quality whereas temporal consistency is measured by TRWC and T-Age. Despite knowing that many existing methods employ DEX that might bias the test results, we opted for DEX due to its accuracy and stability. However, it's important to note that we did not use any age estimation network in our training process, including DEX. 

\begin{table*}[t]
    \caption{Quantitative comparison on CelebV \& VFHQ datasets. The best results are highlighted in bold.}
    \setlength\tabcolsep{4pt}
    \centering
    \begin{tabular}{c c c c c c c c c}
     
    \toprule
    \multirow{2}{*}{Dataset} & \multirow{2}{*}{Models} & \multicolumn{3}{c}{Young $\rightarrow$ Old} & \multicolumn{3}{c}{Old $\rightarrow$ Young} \\
    & & $\text{TRWC}_1$ & T-Age  & & $\text{TRWC}_1$ & T-Age  \\

    \midrule
   
    \multirow{5}{*}{CelebV-HQ} & AgeTransGAN & 4.25 & 1.26 & & 3.12 & 2.29 \\    
    & CUSP & - & - &&  1.90 & 2.19 & \\        
    & FRAN & 1.55 & 1.01 & & 0.74 & \textbf{0.93} &  \\        
    & \textbf{OURS} & \textbf{1.38} & \textbf{0.84}  & & \textbf{0.70} & 1.03 \\   
        
    \midrule
    
    \multirow{5}{*}{VFHQ} & AgeTransGAN & 4.26 & 1.61 & & 2.92 & 2.04  \\
    & CUSP  & - & - && 1.76 & 2.14  \\
    & FRAN & 1.54 & 1.52 & & 0.72 & 1.47  \\    
    & \textbf{OURS} & \textbf{1.35} & \textbf{1.16}& &\textbf{0.69} & \textbf{1.34}  \\
    \bottomrule
    
    \end{tabular}
     \label{tab:video_comparison_switched}
     \vspace{-0.1em}
\end{table*}

\begin{table*}[t]
    \caption{User Study Results. Participants evaluated the methods according to four criteria: Age Accuracy (\textbf{AA}), Identity Preservation 
 (\textbf{IP}), Temporal Consistency (\textbf{TC}), and Overall Naturalness (\textbf{ON}). The highest-scoring results are highlighted in bold. Each score is represented as percentage (\textbf{\%}). }
    \setlength\tabcolsep{4pt}
    \centering
    \begin{tabular}{c | c c c c c | c c c c c c}
     
    \toprule
    \multirow{2}{*}{Models} & \multicolumn{5}{c}{Young $\rightarrow$ Old} & \multicolumn{5}{c}{Old $\rightarrow$ Young} \\
    & AA & IP & TC & ON & Avg& AA & IP & TC & ON& Avg  \\

    \midrule
   
    AgeTransGAN & 7.6 & 9.3 & 7.9 & 9.0 & 8.45 &14.8 & 19.8 & 17.4 & 16.3& 17.08 \\    
    CUSP & 15.7 & \textbf{36.3} & 14.9 & 21.1 & 22.0 &8.3 & 8.9 & 7.6 & 8.3& 8.28 \\        
    FRAN & 28.1 & 21.2 & 29.1 & 27.3 & 26.42 &33.9 & \textbf{37.8} & 35.7 & 35.4& 35.7 \\        
    \textbf{OURS} & \textbf{48.7} & 33.2 & \textbf{48.1} & \textbf{42.6}& \textbf{43.15}& \textbf{43.0} & 33.5 & \textbf{39.3} & \textbf{40.0}& \textbf{38.95}  \\

    \bottomrule
    
    \end{tabular}
     \label{tab:user_studies}
\end{table*}

\section{Comparison Results}
\subsection{Quantitative Results}

\begin{figure}[!t]
\includegraphics[width=\linewidth]{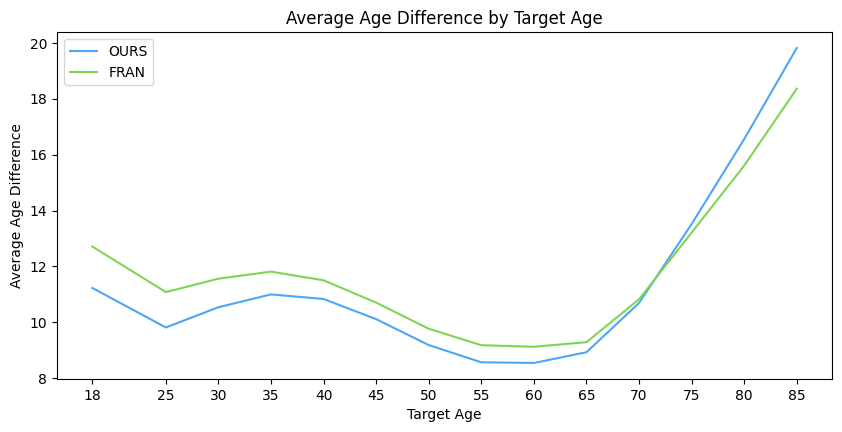} 
\caption{Performance comparison for all target ages. We compare our method with FRAN \cite{zoss2022production} for all target ages.}
\label{fig:all_target_ages}
\end{figure}

We quantitatively compare our method with state-of-the-art methods in Table \ref{tab:video_comparison_switched}. Note that we omit the CUSP result for $Young \rightarrow Old$ due to its maximum target age limitation (65)\cite{gomez2022custom}. Our results indicate that model trained on videos show higher TRWC, suggesting that it maintains greater temporal stability and exhibits lower perceptual differences between the consistent frames. On other side, When evaluating age transformation continuity from a broader perspective using metrics like T-AGE, our method performs comparable or better performance.  These results suggest that these metrics' performances are strongly correlated with the target ages. Therefore, we show the MAE for all the target ages in Fig. \ref{fig:all_target_ages}. We can observe that the age transformation performance of our method is consistently better than the state-of-the-art method \cite{zoss2022production}, despite being trained on synthetic videos. Additionally, our method also improves overall temporal consistency. It is worth mentioning that all the results we presented are in line with user studies (Table. \ref{tab:user_studies}), which further affirm our newly proposed metrics.

\subsection{User Study}

We conducted a subjective evaluation via a user study, presenting 15 videos to a total of 43 anonymous participants for the older direction and 36 participants for the younger direction. This study compares our method with state-of-the-art techniques such as CUSP \cite{gomez2022custom}, AgeTransGAN \cite{hsu2022agetransgan}, and FRAN \cite{zoss2022production} on $Young \rightarrow Old$ and $Old \rightarrow Young$ tasks. 

As illustrated in Table. \ref{tab:user_studies}, a clear majority of participants found our method superior, particularly in terms of age accuracy, temporal consistency, and naturalness. These findings highlight the effectiveness of our video-based approach in enhancing both temporal consistency and age transformation capabilities, thereby demonstrating that proposed metrics are reliable indicators of cognitive temporal consistency. For example, the significance of a 0.18 (±1) difference in TRWC in Table. \ref{tab:video_comparison_switched} is evident by the users' choices.

\begin{figure*}[!t]
  \centering
  \includegraphics[width=\textwidth]{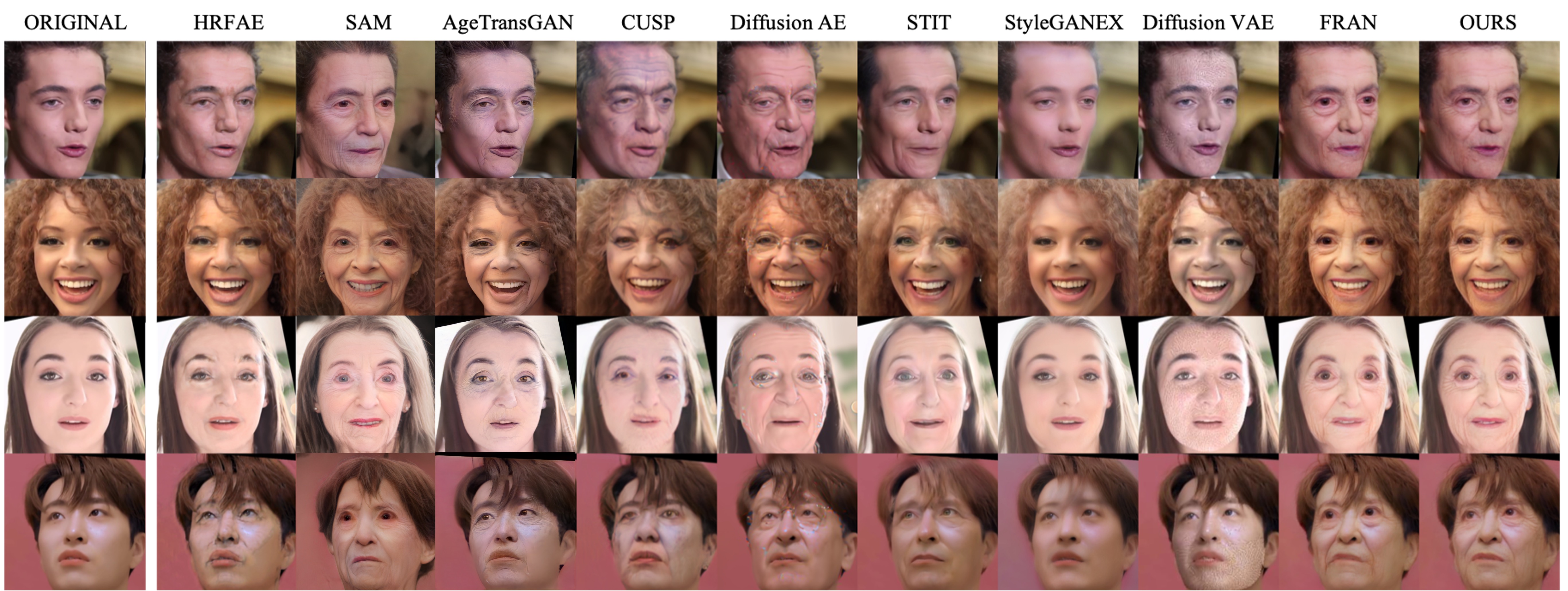}
  \caption{Qualitative comparison with existing state-of-the-art methods. The target age is set to 85.}
  \label{qualitative_results}
\end{figure*}

\subsection{Qualitative Results}

We present our qualitative comparison with the existing state-of-the-art in Fig. \ref{qualitative_results}. It is evident that SAM is effective in age transformation, while STIT and StyleGANEX struggle in this aspect. However, SAM fails to preserve attributes such as identity, pose, and expression because it strongly depends on age classifier loss, which is proficient at altering age but not at maintaining other key attributes. On the other hand, STIT and StyleGANEX, which utilize average vectors related to age changes, fail to properly consider the individual samples. As a result, they encounter difficulties in age transformation, particularly when dealing with wild samples. In such cases, their methods become more susceptible due to their strong reliance on trained data for constructing the latent space. Diffusion AE achieves successful age transformation but introduces artifacts such as glasses, leading to identity drift. While the Diffusion VAE works for small changes, it fails entirely in age transformation when dealing with a significant age gap for all subjects.Image-based methodologies, such as HRFAE and AgeTransGAN, inherently exhibit  lower age transformation capabilities in wild scenarios. In contrast, while CUSP achieves successful age transformation in certain samples, it often tends to produce artifacts. In contrast, our method successfully performs age transformation while maintaining stable results in terms of identity, expression, and pose. While FRAN also achieves similar quality, our results possess more natural wrinkle details, with the differences being particularly pronounced around the eyes. 

\begin{figure*}[!t]
\includegraphics[width=\textwidth]{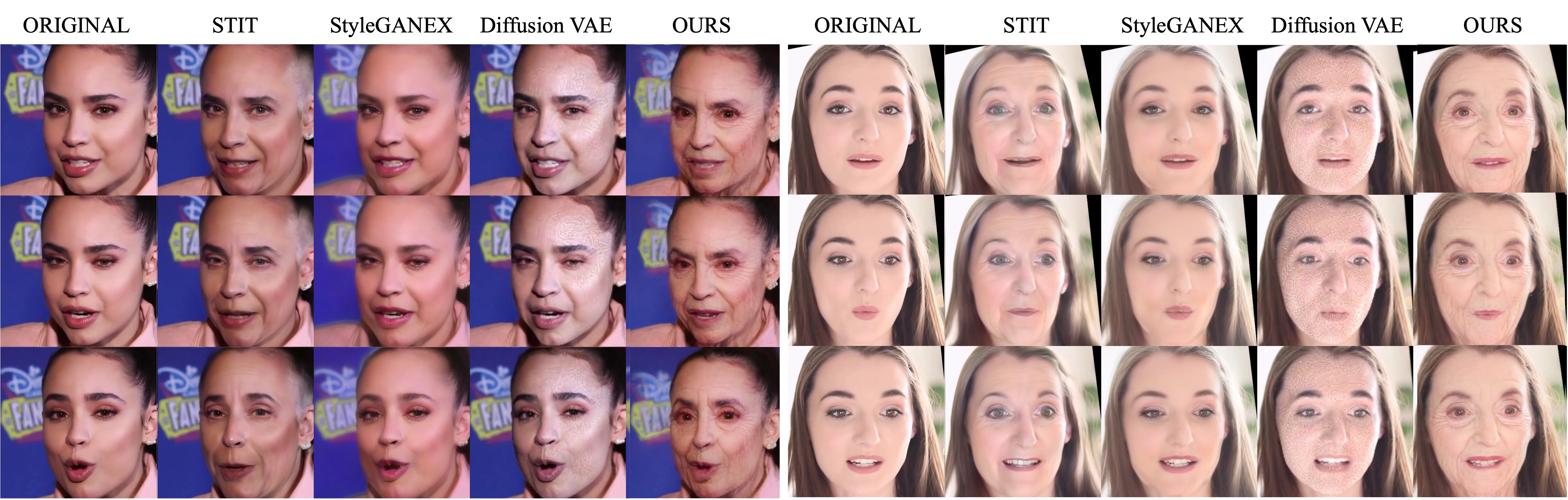} 
\caption{Comparison with video editing methods.}
\label{fig:temporal_comparison}
\end{figure*}

Furthermore, Fig. \ref{fig:temporal_comparison} presents a performance comparison of different frames to demonstrate the efficacy of our approach against other video editing methods. It can be observed that, despite producing temporally consistent results, their ability to transform age is quite limited. Overall, these comparison results indicate that leveraging a video dataset enables our method to outperform the existing methods, yielding natural age progression and high-quality results, even under extreme conditions. Similar to every method, our method has some limitations which are discussed in Sec. \ref{sec:limitations_supp}, along with possible future directions. More results can be found in Sec. \ref{sec:comparison_supp}.

\section{Ablation Study}
In this section, we ablate the selection of the modules of proposed pipeline except the choice of SAM \cite{alaluf2021only} which is already validated in \cite{zoss2022production}.

\subsubsection{Key Frame Generation}

\begin{table*}[t]
\centering
\caption{Ablation studies: We ablate the (a) key frame generation and (b) different frame interpolation methods to select the highest performance method based on higher CPBD and lower warping error.}
\begin{tabular}{cc}%
\begin{minipage}[t]{0.5\textwidth}
\centering
\caption*{(a) Key frame generation}
\begin{tabular}{ccccc}
\toprule
Method & \multicolumn{2}{c}{MAE ($\downarrow$)} & \multicolumn{2}{c}{ID ($\uparrow$)} \\ 
       & 18       & 85         & 18         & 85           \\ 
\midrule
StyleHEAT & 8.06 & 4.41 & 0.64 & 0.50 \\
StyleMask & 7.23 & 9.98 & 0.61 & 0.37 \\ 
OSFV      & \textbf{1.62} & \textbf{0.71} & \textbf{0.90} & \textbf{0.90} \\
\bottomrule
\end{tabular}
\end{minipage}
&
\begin{minipage}[t]{0.5\textwidth}
\centering
\caption*{(b) Motion generation}
\begin{tabular}{lccc}
\toprule
Methods & DQBC & EMA-VFI & FILM\\
\midrule
CPBD $\uparrow$ & 0.5115 & 0.4213 & \textbf{0.6061} \\
Warp Error $\downarrow$ & 0.00118 & 0.00119 & \textbf{0.00104}\\
\bottomrule
\end{tabular}
\end{minipage}%
\end{tabular}
\label{tab:ablation_studies}
\end{table*}

In Table \ref{tab:ablation_studies} (a), we compare the performance of three reenactment methods: OSFV \cite{wang2021one}, StyleMask \cite{bounareli2023stylemask}, and StyleHeat \cite{yin2022styleheat}, for key frame generation. The best method is selected for our re-aging task based on lower MAE and higher identity similarity (ID). For ID, we calculate the cosine similarity between the vectors of the face image, extracted using ArcFace \cite{deng2019arcface}. Given different pose and expressions, it is crucial to maintain the age of a person without losing its identity. The results show that OSFV \cite{wang2021one} successfully preserves both the identity and age of the person, while other methods fail to maintain the person's identity and age. We further investigate the training configurations of \cite{wang2021one} in Sec. \ref{sec:training_configuration_supp}.

\subsubsection{Motion Generation}

In this ablation study, we experiment with state-of-the-art frame interpolation models \cite{reda2022film,zhang2023extracting,zhou2023video} to determine the most effective motion generation method. We evaluate image quality and motion consistency using non-reference-based metrics such as CPBD \cite{narvekar2009no, narvekar2011no} for image quality and warping error \cite{zhang2023metaportrait} for motion consistency. Table \ref{tab:ablation_studies} (b) suggests that all methods exhibits lower error in terms of temporal consistency. However, FILM \cite{reda2022film} outperforms other models by producing sharp results, while the outputs of the remaining models tend to be blurry.

\begin{table}[t]
\caption{Ablating our method with image (I) and video (V) discriminators on VFHQ dataset.}
\centering
\setlength{\tabcolsep}{2pt} 
\begin{tabular} {c  c c c }

\toprule
 Task &  Method &  {$\text{TRWC}_1$} $\downarrow$   &  MAE $\downarrow$ \\

\cline{1-4}

\multirow{2}{*}{Old$\rightarrow$Young}

&  I  & 0.72 & 10.7\\ 

& I + V & {\bf0.69} & {\bf 10.6} \\

\cline{1-4}
\multirow{2}{*}{Young$\rightarrow$Old}

& I  & 1.53 & {\bf16.9} \\\

& I+V & {\bf1.35} & 19.7 \\
\bottomrule

\end{tabular}
\label{tab:ablation_video_discriminator}
\end{table}

\subsubsection{Video Discriminator}

In Table \ref{tab:ablation_video_discriminator}, we present ablation studies to assess the effect of the video discriminator. These results indicate that incorporating both image and video discriminators consistently demonstrates superior performance as compared to solely relying on the $I$ discriminator. Whereas, for the Old to Young task, the difference is marginal; this is attributed to the fact that the consistency of wrinkles is often consistent in younger images. Overall, the results suggest that using two discriminators, I + V, improves the consistency of generated videos with negligible effect on MAE.

\section{Ethical statement}
\label{sec:ethics_supp}

Our proposed video dataset comprises synthetic videos which heavily relies on StyleGAN images, trained on FFHQ dataset \cite{karras2019style}. We acknowledge the potential biases inherited from StyleGAN and FFHQ. Recognizing the societal threat or risk of misuse of our work, we explicitly disapprove of any malicious applications of our research. Our primary intent is to contribute positively for the production and advertisement industry. 

\section{Conclusion}
\label{sec:conclusion}

In this paper, we introduced a new video dataset tailored for the supervised learning of video re-aging. This dataset encompasses subjects from a wide range of age groups. Shifting from the conventional model-centric focus, we adopted a data-centric approach. Inspired by recent advancements in face reenactment and frame interpolation, we encompasses various facial poses and expressions in the proposed data. Based on this dataset, we demonstrated the capabilities of the proposed method, which features a baseline network architecture that emphasize both temporal consistency and the quality of age transformations. Furthermore, we formulated two novel metrics to evaluate temporal consistency in video re-aging, which consider age-relevant features such as facial wrinkles over time. We validated our method with comprehensive experiments.

\bibliographystyle{splncs04}
\bibliography{main}

\clearpage
\setcounter{page}{1}

\noindent \subsubsection*{\textcolor{Peach}{Referring to the main article, Line Number is specified}}

\section{Experimental Settings}
\label{sec:Experimental}
\subsection{Network Architecture}
\noindent\textcolor{Peach}{(Line 204-205: Please refer to Sec. 9 for the details of U-Net network architecture.)}
\vspace{-2em}
\subsubsection{Generator}

Our network generator consists of a shared recurrent block, which is essentially a U-Net-based network which comprises both up-sampling and down-sampling layers. For every set of 3 input frames, we mask the frames using input and target ages, thereby increasing the number of input channels from 3 to 5. Each of these input frames has a resolution of $512\times512$. In the first layer of the recurrent block, we concatenate three input frames with a 64-channel hidden state and a 3-channel output image from the previous iteration. This combination results in a spatial image with a total of 82 channels. Such a detailed configuration ensures that the spatial image accurately captures both the current input data and the information processed in the preceding steps. Finally, we obtain a 67-channel spatial image that is subsequently split into a delta image and a hidden state. For detailed architectures of these layers and blocks, please refer to Table \ref{tab:downsample} for down-sampling, Table \ref{tab:upsample} for up-sampling, Table \ref{tab:recurrent_block} for the recurrent block, and Table \ref{tab:generator} for generator architecture.

\begin{table}[!ht]    
\caption{Down-sampling layer.}
    \centering
    \begin{tabular}{l r}   
    \hline 
        Layer & Output \\ \hline 
        Input & $w \times h \times c$ \\ \hline
        MaxBlurPool & $w/2 \times h/2 \times c$ \\ 
        $3\times3$ Conv  + LeakyReLU  & $w/2 \times h/2 \times 2c$ \\ 
        $3\times3$ Conv  + LeakyReLU  & $w/2 \times h/2 \times 2c$ \\     \hline    
        Output & $w/2 \times h/2 \times 2c$ \\ \hline
    \end{tabular}

\label{tab:downsample}
\end{table}

\begin{table}[!ht]     
\caption{Up-sampling layer.}
    \centering
    \begin{tabular}{l r}   
    \hline
        Layer & Output \\ \hline 
        Input & $w \times h \times c$ \\ \hline
        BlurUpSample & $2w \times 2h \times c$ \\ 
        $3\times3$ Conv  + LeakyReLU  & $2w \times 2h \times c/2$ \\ 
        $3\times3$ Conv  + LeakyReLU  & $2w \times 2h \times c/2$ \\     \hline    
        Output & $2w \times 2h \times c/2$ \\ \hline
    \end{tabular}

\label{tab:upsample}
\end{table}

\begin{table}[!ht]
\caption{Recurrent block.}
    \centering
    \begin{tabular}{l r}   
    \hline
        Layer                         & Output \\ \hline 
        Input (Video)                         & $512 \times 512 \times 3  \times 5$ \\    
        Reshape                       & $512 \times 512 \times 15$ \\     
        Previous Hidden State                  & $512 \times 512 \times 64$ \\     
        Previous Output               & $512 \times 512 \times 3$ \\         
        Concatenation                 & $512 \times 512 \times 82$ \\ \hline        
        $3\times3$ Conv  + LeakyReLU  & $512 \times 512 \times 64$ \\ 
        $3\times3$ Conv  + LeakyReLU  & $512 \times 512 \times 64$ \\ 
        DownSampleLayer               & $256 \times 256 \times 128$ \\ 
        DownSampleLayer               & $128 \times 128 \times 256$ \\ 
        DownSampleLayer               & $64 \times 64 \times 512$ \\ 
        DownSampleLayer               & $32\times 32 \times 1024$ \\ \hline
        UpSampleLayer                 & $64 \times 64 \times 512$ \\ 
        UpSampleLayer                 & $128 \times 128 \times 256$ \\ 
        UpSampleLayer                 & $256 \times 256 \times 128$ \\ 
        UpSampleLayer                 & $512 \times 512 \times 64$ \\ 
        $1\times1$ Conv               & $512 \times 512 \times 67$ \\     \hline    
        Output Delta Image            & $512 \times 512 \times 3$  \\ 
        Output Hidden State + LeakyReLU                  & $512 \times 512 \times 64$  \\ \hline
    \end{tabular}

\label{tab:recurrent_block}
\end{table}

\begin{table}[!ht]
\caption{Generator architecture. $N$ represents the number of frames in the input sequence.}
    \centering
    \begin{tabular}{l r}   
    \hline
        Layer            & Output \\ \hline 
        Input (Video)    & $ 512 \times 512 \times N \times 3 $ \\ 
        Recurrent Blocks ($\times$ N) & $  512 \times 512 \times N \times 67$ \\ \hline 
        Output (Video)   & $ 512 \times 512 \times N  \times 3 $ \\ \hline 
    \end{tabular}

\label{tab:generator}
\end{table}

\subsubsection{Discriminator}

In our architecture, we use two discriminators. The first discriminator, specifically focused on image quality, is based on PatchGAN \cite{pix2pix2017}. This consists of three downsampling layers followed by two convolution layers. The output frames are  concatenated by their target age masks in a channel-wise manner, resulting in a 4-channel input image. Additionally, we process all output frames independently by concatenating them batch-wise as shown in Table \ref{tab:discriminator_architecture}. 

We also incorporate another discriminator with 3D convolutions, referred to as the video discriminator, to evaluate the realism of motion. Our video discriminator consists of three downsampling layers. The input is created by concatenating the target input age mask and the output images corresponding to three consecutive frames, resulting in a four-channel input. The architecture of the video discriminator is illustrated in Table \ref{tab:video_discriminator_architecture}. It is noted that both of our discriminators consist of LeakyReLU in every layer, which is not shown in Table \ref{tab:discriminator_architecture} and Table \ref{tab:video_discriminator_architecture}.

\begin{table}[!ht]
\caption{Architecture of image discriminator.}
    \centering
    \begin{tabular}{l r}   
    \hline
        Layer                         & Output \\ \hline 
        Video with Target Mask       & $  512 \times 512 \times 4 $ \\
        $4\times4$ Conv  & $ 256 \times 256 \times 64$ \\     
        $4\times4$ Conv  & $ 128 \times 128 \times 128$ \\     
        $4\times4$ Conv  & $64 \times 64 \times 256$ \\     
        $4\times4$ Conv (Stride = 1)  & $64 \times 64 \times 512$ \\     
        $4\times4$ Conv (Stride = 1)  & $64 \times 64 \times 1$ \\  \hline   
    \end{tabular}
    
    \label{tab:discriminator_architecture}
\end{table}

\begin{table}[!ht]
\caption{Architecture of video discriminator.}
    \centering
    \begin{tabular}{l r}   
    \hline
        Layer                         & Output \\ \hline 
        Video with Target Mask       & $  512 \times 512 \times 3 \times 4 $ \\
        $4\times4$ 3D Conv   & $ 256 \times 256 \times 32 \times 4$  \\     
        $4\times4$ 3D Conv   & $ 128 \times 128 \times 64 \times 4$ \\     
        $4\times4$ 3D Conv   & $64 \times 64 \times 128 \times 4$ \\     
        $4\times4$ 3D Conv (Stride = 1)   & $64 \times 64 \times 256 \times 4$ \\     
        $4\times4$ 3D Conv (Stride = 1)   & $64 \times 64 \times 1 \times 4$ \\  \hline   
    \end{tabular}
    
    \label{tab:video_discriminator_architecture}
\end{table}

\subsection{Implementation settings}
\label{sec:implementation_settings_supp} 
\noindent \textcolor{Peach}{(Line 283-284: We provide the additional details in the supplementary document)}

\begin{figure*}[!htbp]
\includegraphics[width=\textwidth, height=\textheight]{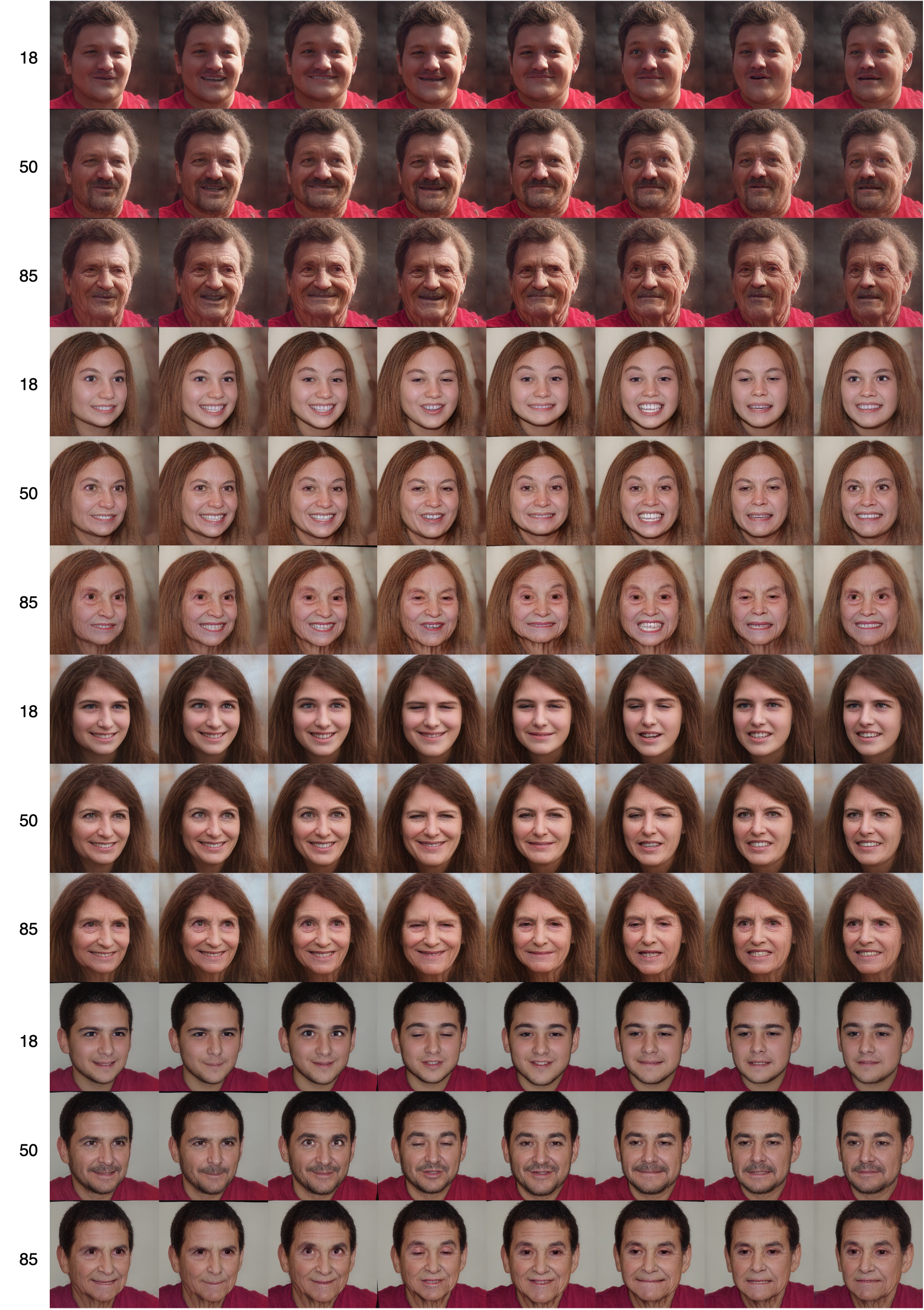} 
\caption{Data samples in our proposed video dataset with ages 18, 50, and 85. }
\label{fig:more_samples_1}
\end{figure*}

In this section, we describe our experimental setup for video re-aging training data. We utilized a total of 4,248 subjects to train our network, generating 14 videos per subject. These subjects were divided into 14 classes, covering a wide age range from 18 to 85. Each video consists of 57 frames for training. We applied a cumulative probability of blur detection(CPBD) threshold of 0.5 to ensure sharpness, reducing the prevalence of blurry videos, especially in those with lower CPBD values. Higher CPBD videos, exhibiting fewer dynamic poses, led us to maintain this threshold to balance sharpness and dynamic representation. The sample images of our generated dataset are shown in Fig. \ref{fig:more_samples_1}. For testing, we selected 20 videos from VFHQ \cite{xie2022vfhq} and 85 videos from CelebV-HQ \cite{zhu2022celebvhq} for each Young $\rightarrow$ Old direction (target ages: 65, 75, 85) and Old $\rightarrow$ Young direction (target ages: 18, 25, 35).

During training, we set input and output ages randomly without imposing any conditions that both ages cannot be equal in the same iteration. This approach enables the reconstruction of the input image when the input and target ages are the same. We used a learning rate of 0.0001 for 250K iterations with a batch size of 4. Additionally, we introduced temporal augmentation in our training. We randomly selected a frame interval $\Delta{t}$ from $[3,5,7]$. Reverse augmentation is applied to every frame sequence with a 0.5 probability. We implemented our code in the PyTorch framework and trained our model with a single A100 GPU.

\subsubsection*{Experimental Settings for Ablation Study}

For the ablation study on reenactment methods in Table \ref{tab:ablation_studies} (a), we employed 5 videos of 10 subjects, each with 2 target ages (18 and 85), totaling 100 videos. Table \ref{tab:ablation_studies} (b), which shows the different interpolation methods, utilizes 21 videos from \cite{xie2022vfhq} for testing.  In the experiments exploring various training configurations of OSFV \cite{wang2021one} (Table \ref{tab:ablation_different_dataset}), we use a relatively large test set comprising 100 videos, as this step is crucial for generating quality data. Lastly, Our ablation study for discriminators in Table \ref{tab:ablation_video_discriminator} follows the same settings as presented in Table \ref{tab:video_comparison_switched}. 

\section{Additional Qualitative Comparison}
\label{sec:comparison_supp}

In this section, we provide additional comparison results to evaluate our method against state-of-the-art approaches.

\begin{figure*}[!htbp]
\includegraphics[width=\textwidth]{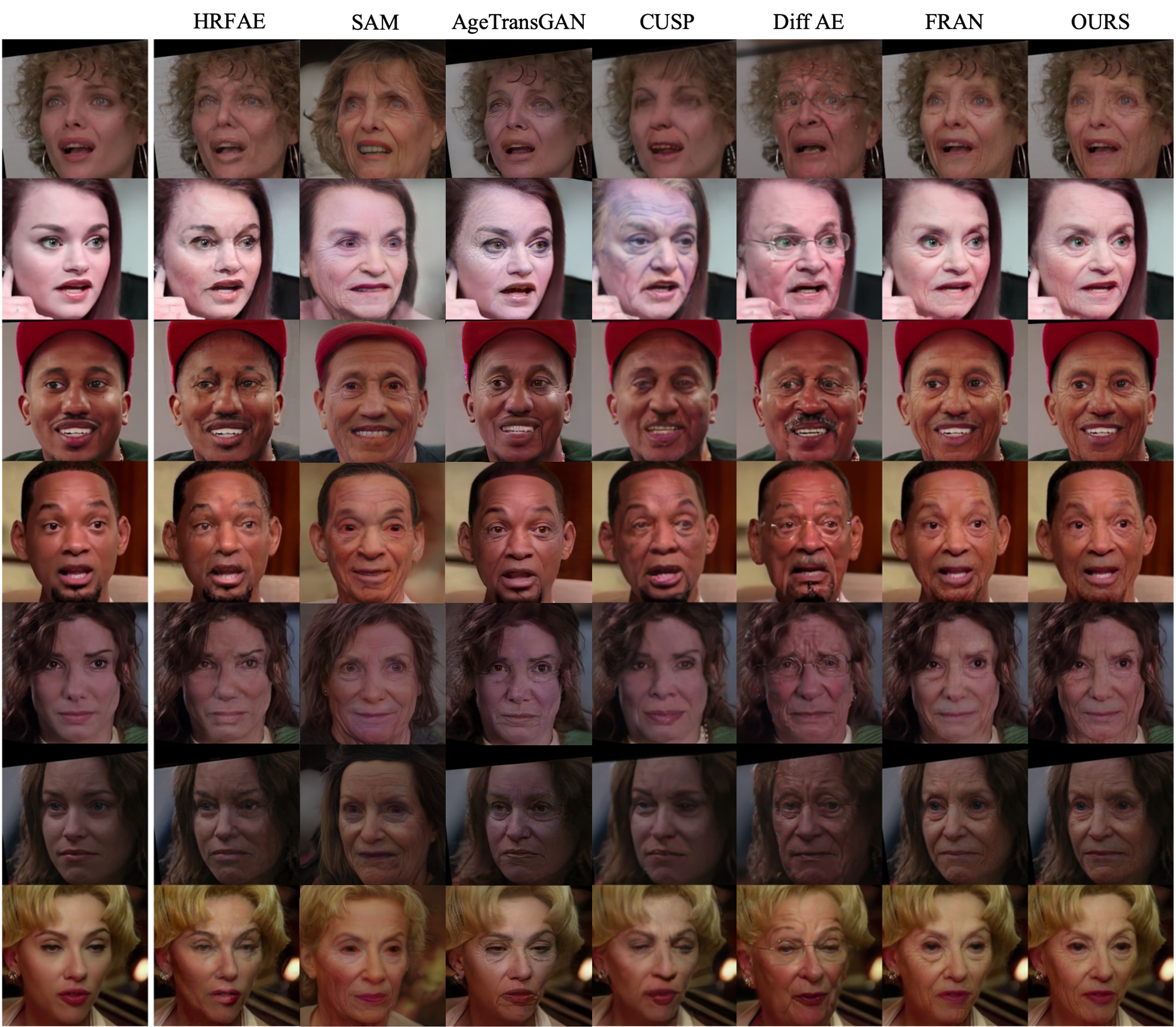} 
\caption{Qualitative comparison with existing state-of-the-art methods on CelebV-HQ dataset. The target age is set to 85.}
\label{fig:fig_SM1_to_old_01}
\end{figure*}

\noindent \textbf{Young $\rightarrow$ Old.} Fig. \ref{fig:fig_SM1_to_old_01} shows the comparison results for $Young \rightarrow Old$. The results indicate that SAM \cite{alaluf2021only} emphasizes on target age and does not retrain the image fidelity especially for side-poses. This tendency is also observed for HRFAE \cite{yao2021high}. The diffusion based model DIFF-AE \cite{preechakul2022diffusion} and CUSP \cite{gomez2022custom} often produce artifacts in output images. Recently proposed AgeTransGAN \cite{hsu2022agetransgan} generate artistic images that appears to be unnatural. We observe that, compared to FRAN \cite{zoss2022production}, our method is robust in handling profile shots and side-view angles.

\begin{figure*}[!htbp]
\includegraphics[width=\textwidth]{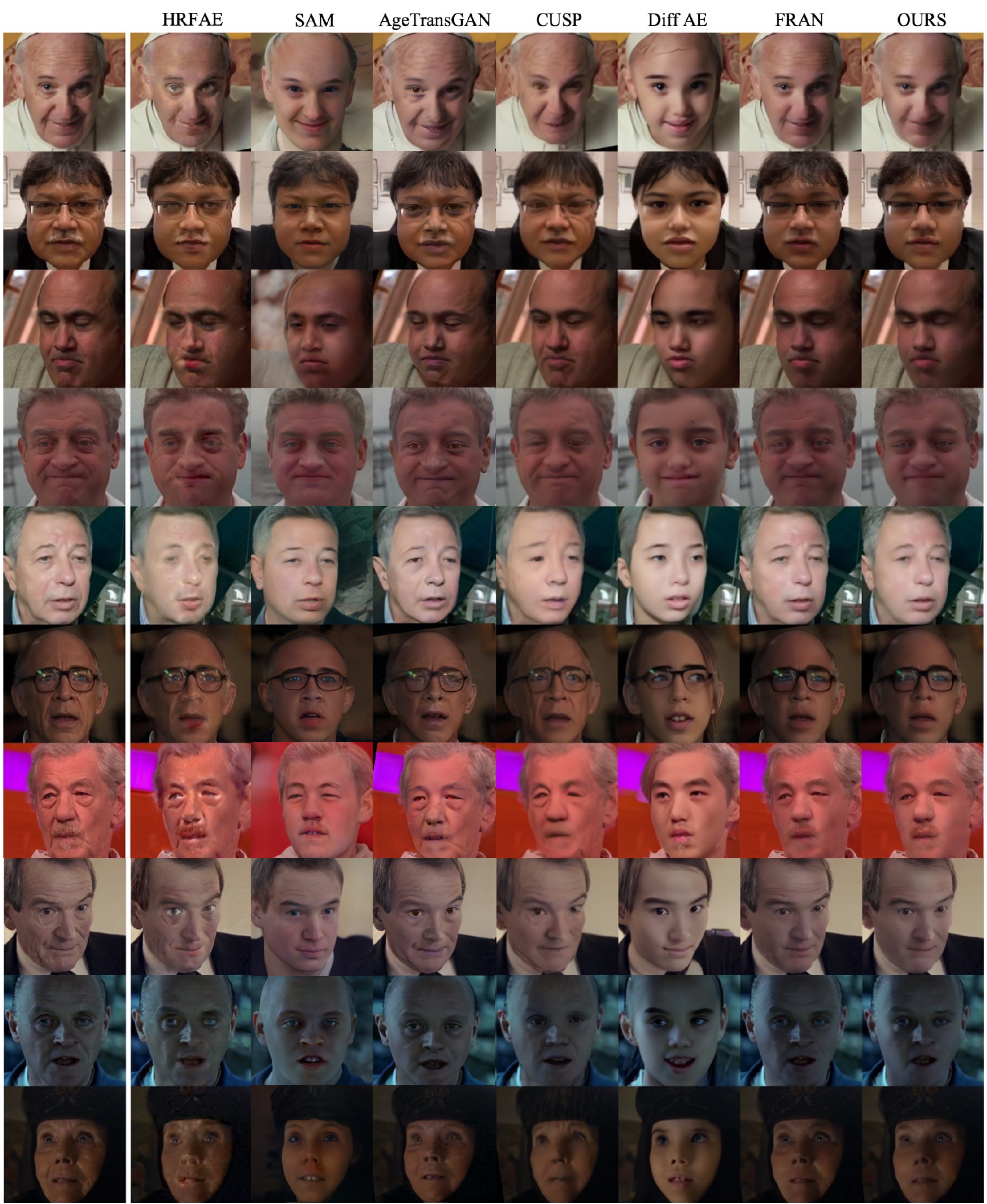} 
\caption{Qualitative comparison with existing state-of-the-art methods on CelebV-HQ dataset. The target age is set to 18.}
\label{fig:fig_SM1_to_young_01}
\end{figure*}

\vspace{1em}
\noindent \textbf{Old $\rightarrow$ Young.} We also present our comparison results for $Old \rightarrow Young$ task in Fig. \ref{fig:fig_SM1_to_young_01}. These results show the similar tendency in which HRFAE \cite{yao2021high} and SAM \cite{alaluf2021only} fail to recover the facial details and suffer from significant artifacts. The results indicate that \cite{preechakul2022diffusion} lacks control on target ages and consists various artifacts in their results with variant ages. \cite{zoss2022production} focuses solely on wrinkles removal  whereas our method generates smoother images, appearing more natural. In addition, it is quite evident that \cite{zoss2022production} often fail to remove forehead wrinkles and Nasolabial folds. 

\begin{figure*}[!htbp]
\includegraphics[width=\textwidth]{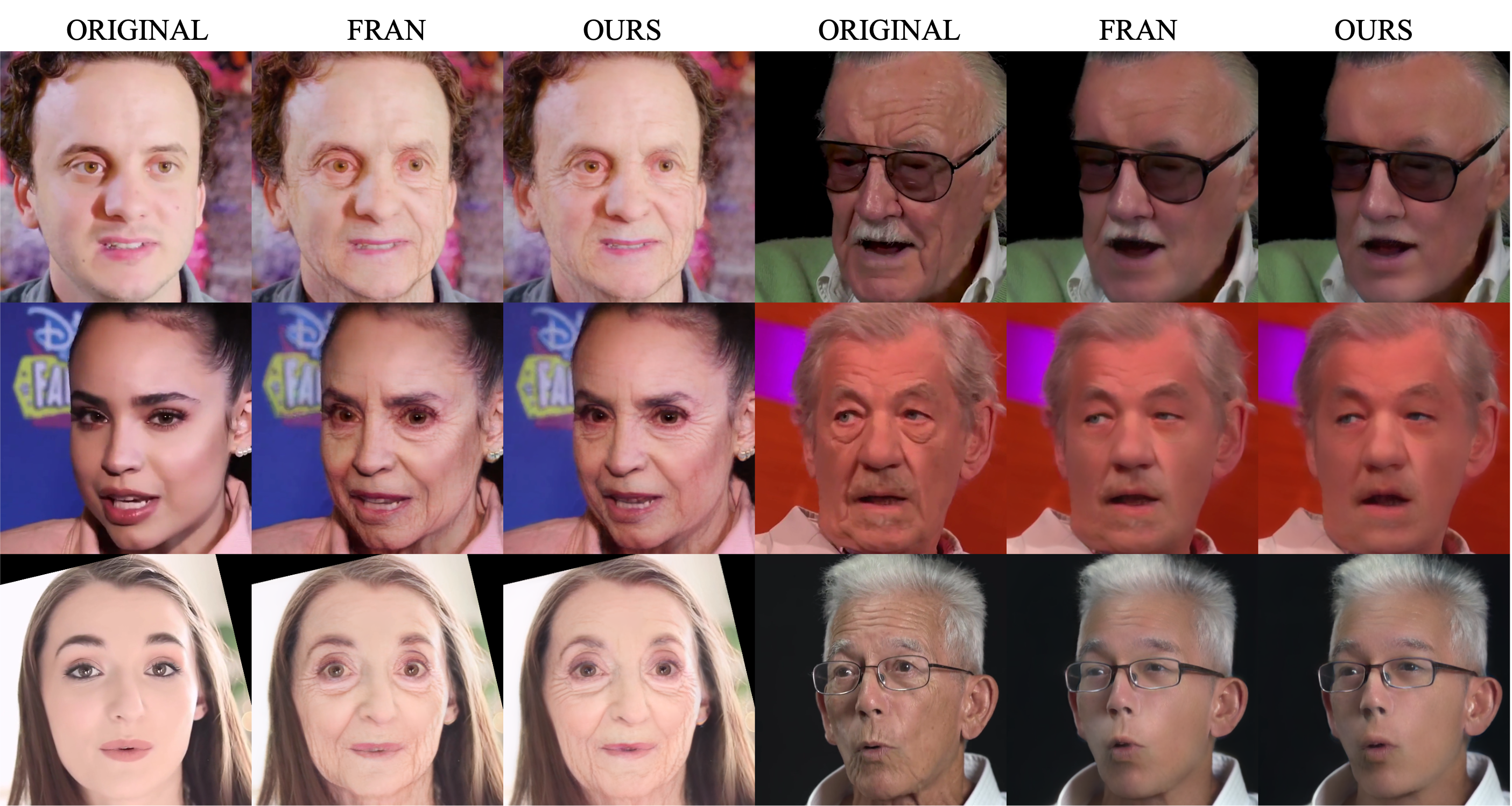} 
\caption{Detailed comparison between FRAN and OURS. Left: $Young \rightarrow Old$, Right: $Old \rightarrow Young$.}
\label{fig:FRAN_vs_VFRAN}
\end{figure*}

\vspace{1em}
\noindent \textbf{Comparison to FRAN.} In Fig. \ref{fig:FRAN_vs_VFRAN}, We conduct a more detailed comparison with FRAN using enlarged images. The rows represent each subject, and the left nine images and right nine images depict Young $\rightarrow$ Old and Old $\rightarrow$ Young transformations, respectively. The results show that our method consistently performs better in age transformation while maintaining natural outcomes across all results. In the Old $\rightarrow$ Young category, it can be observed that the wrinkles around the eyes are unnaturally generated in the results produced by FRAN. Particularly in the transition to a younger appearance, it is noticeable that wrinkles on the forehead are removed more effectively with our method compared to FRAN.

\section{Ablation of Training Configurations of OSFV}
\label{sec:training_configuration_supp}

\noindent\textcolor{Peach}{(Line 162: (Footnote) Additional details are provided in the supplementary materials.)}
For  generation of key frames, we utilize the unofficial implementation\footnote{\href{https://github.com/zhengkw18/face-vid2vid}{https://github.com/zhengkw18/face-vid2vid}} of OSFV \cite{wang2021one} trained on VoxCeleb \cite{nagrani2017voxceleb} at 256$^2$ resolution. We denote this configuration as model `\textbf{A}'. We evaluated the performance on 100 \cite{xie2022vfhq}'s test set For the training at 512$^2$, we experimented with four different training configurations, as detailed in Table \ref{tab:ablation_different_dataset}. One of the approaches involved a naive upscaling of the images to 512$^2$, resulting in a notably poor CPBD. Afterward, we fine-tuned the existing pretrained model on VFHQ \cite{xie2022vfhq} at 512$^2$, resulting in improved quality and designated as model `\textbf{B}'. We also conducted training from scratch at $256^2$ on the VFHQ dataset, following \cite{wang2021one}. While SSIM and PNSR showed improvement compared to model `\textbf{A}', CPBD remained lower. This model is referred to as model `\textbf{C}'. Therefore, we fine-tuned model `\textbf{C}' at $512^2$ and observed an overall improvement in the dataset's quality, designated as model `\textbf{D}'. For the test set, we randomly selected 100 videos from \cite{xie2022vfhq}'s test set. Our configurations are summarized as follows:

\begin{tabbing}
\textbf{A:}  Publicly available model trained at resolution 256$^2$ \\
\textbf{B:}  Fine-tuning model A on VFHQ at resolution 512$^2$ \\
\textbf{C:}  Training \cite{wang2021one} from scratch on VFHQ at resolution 256$^2$ \\
\textbf{D:}  Fine-tuning C on VFHQ at resolution 512$^2$ \\
\end{tabbing}

\begin{table}[!h]
\caption{ Ablating performance of OSFV \cite{wang2021one} with different training configurations. }

\small
\centering

\setlength{\tabcolsep}{0.8em}{
\begin{tabular}{cccc}
    \toprule
    Method & PSNR$\uparrow$  & SSIM$\uparrow$  & CPBD$\uparrow$  \\
    \midrule
    \textbf{A}  & 17.553 & 0.657 & 0.149 \\
    \textbf{B}  & 18.476 & 0.665 & 0.442 \\
    \textbf{C}  & 19.200 & \textbf{0.697} & 0.223 \\
    \textbf{D}   & \textbf{19.519} & 0.683 & \textbf{0.487} \\
    \bottomrule
\end{tabular}
}
\label{tab:ablation_different_dataset}
\end{table}

\section{Limitations and Future Works}
\label{sec:limitations_supp}
\noindent\textcolor{Peach}{(Line 354-355: our method has some limitations \dots)}
\noindent\textcolor{Peach}{(Line 368-369: We further investigate the training configurations \dots)}

\begin{figure}[!h]
\centering
\includegraphics[width=0.5\textwidth, height=0.5\textheight]{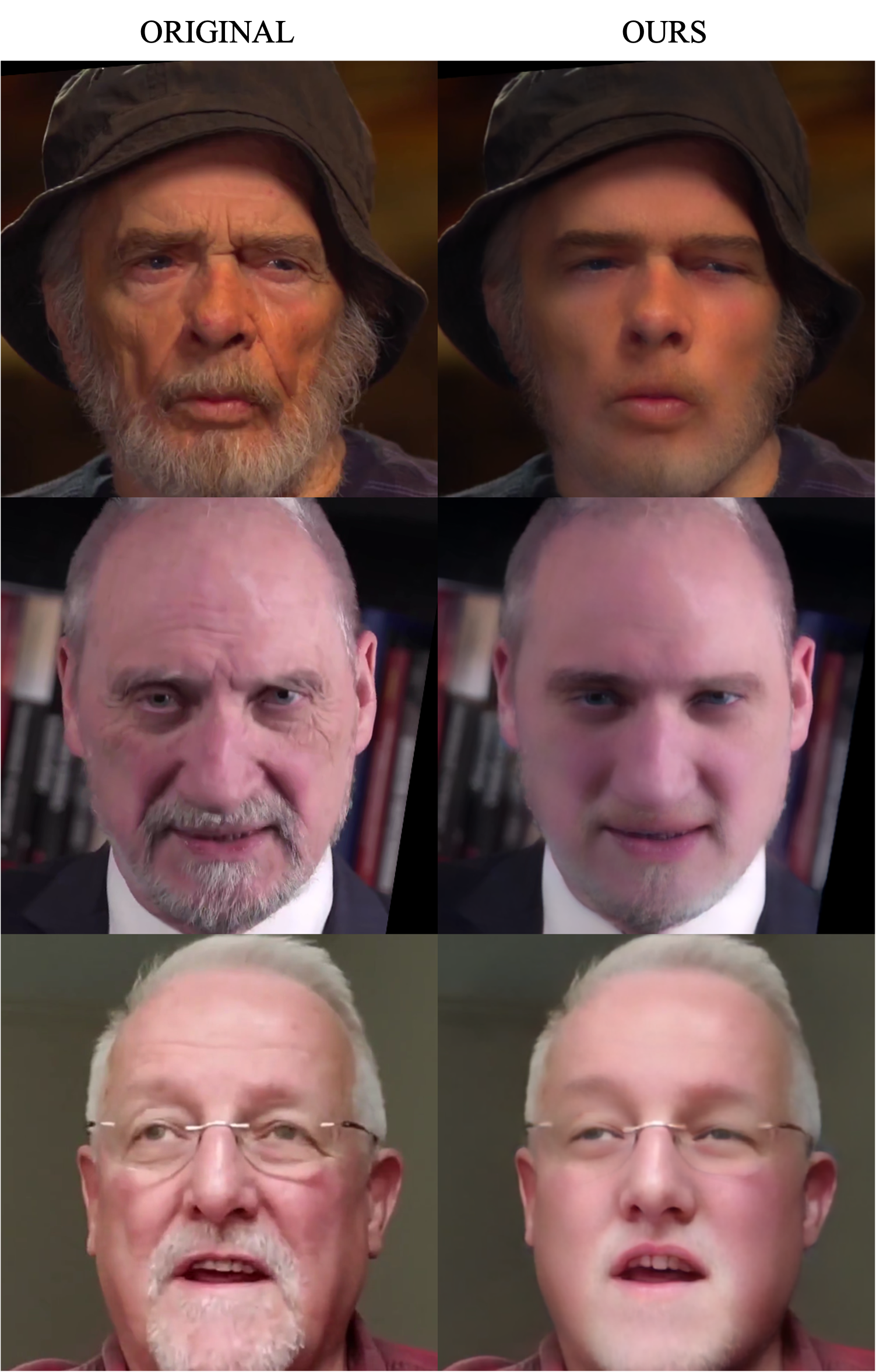} 
\caption{Limitations of our approach.} 
\label{fig:limitation}
\end{figure}

Despite of our work surpasses current state-of-the-art methods, we have observed that our method often struggles to preserve the facial hairs. The limitation of our methods are shown in Fig. \ref{fig:limitation}. Our empirical results found that this problem arises due to SAM \cite{alaluf2021only}. Therefore, leveraging alternative methodologies that are closely aligned to our aspirations may improve the performance. Additionally, we employ simple encoder-decoder within recurrent block architecture. One can explore more advanced network architectures, focusing on refining the transformation of facial shapes. This will likely lead to enhancements in age transformation capabilities and temporal consistency. Furthermore, we have utilized \cite{wang2021one} for face reenactment and \cite{reda2022film} for frame interpolation, integrating models such as \cite{zhang2023metaportrait} could potentially lead to the creation of even more realistic videos and enhancing our performance. These tasks require further investigation by future researchers.

\end{document}